\documentclass[10pt,twocolumn,letterpaper]{article}

\usepackage[accsupp]{axessibility}
\usepackage{ijcb}
\usepackage{times}
\usepackage{epsfig}
\usepackage{graphicx}
\usepackage{amsmath}
\usepackage{amssymb}
\usepackage{multirow, booktabs}

\ijcbfinalcopy 

\ifijcbfinal\fi
\begin{document}

\title{Collaborative Feature Learning for Fine-grained Facial Forgery\\ Detection and Segmentation}


\author{
    Weinan Guan\textsuperscript{\rm 1, 2}, 
    Wei Wang\textsuperscript{\rm 2}\thanks{Corresponding Author}, 
    Jing Dong\textsuperscript{\rm 2}, 
    Bo Peng\textsuperscript{\rm 2},  
    Tieniu Tan\textsuperscript{\rm 2}\\
    \textsuperscript{\rm 1}School of Artificial Intelligence, University of Chinese Academy of Sciences\\
    \textsuperscript{\rm 2}Center for Research on Intelligent Perception and Computing, CASIA\\
    {\tt\small\texttt{weinan.guan@cripac.ia.ac.cn,\{wwang,jdong,bo.peng,tnt\}@nlpr.ia.ac.cn}}
}

\maketitle
\thispagestyle{empty}

\begin{abstract}
   Detecting maliciously falsified facial images and videos has attracted extensive attention from digital-forensics and computer-vision communities. An important topic in manipulation detection is the localization of the fake regions. Previous work related to forgery detection mostly focuses on the entire faces. However, recent forgery methods have developed to edit important facial components while maintaining others unchanged. This drives us to not only focus on the forgery detection but also fine-grained falsified region segmentation. In this paper, we propose a collaborative feature learning approach to simultaneously detect manipulation and segment the falsified components. With the collaborative manner, detection and segmentation can boost each other efficiently. To enable our study of forgery detection and segmentation, we build a facial forgery dataset consisting of both entire and partial face forgeries with their pixel-level manipulation ground-truth. Experiment results have justified the mutual promotion between forgery detection and manipulated region segmentation. The overall performance of the proposed approach is better than the state-of-the-art detection or segmentation approaches. The visualization results have shown that our proposed model always captures the artifacts on facial regions,  which is more reasonable.
\end{abstract}

\section{Introduction}
    Recently, with the rapid development of deep learning, especially Generative Adversarial Networks (GAN) \cite{NIPS2014_5423}, the techniques of artificial content generation have made impressive advances. Among these artifacts, the maliciously manipulated face images and videos, which are commonly recognised as deepfakes, can easily deceive human and mislead the public opinions. Therefore, deepfake detection  attracts extensive attention worldwide.
    
    \begin{figure}[htb]
    	\centering
    	\includegraphics[width=0.45\textwidth]{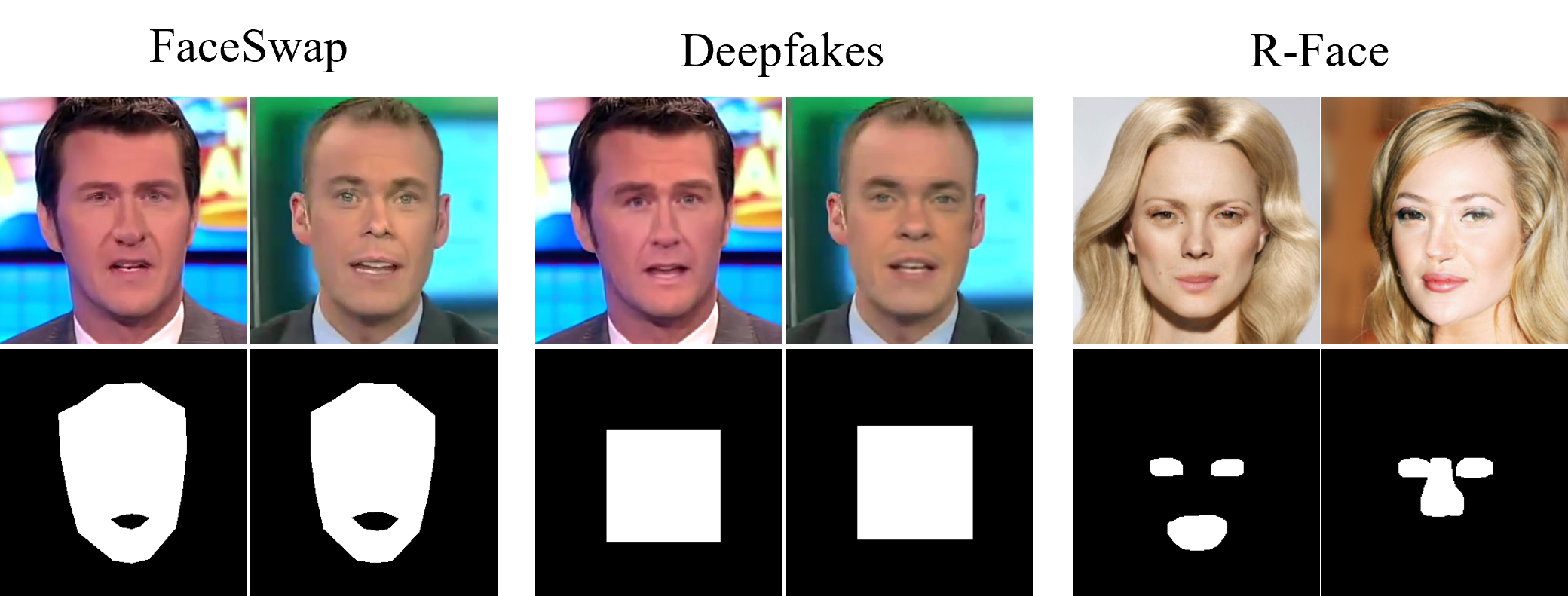}
    	\caption{Example faces and manipulation masks in our dataset.}
    	\label{fig:dataset-examples}
    \end{figure}        
    Consequentially, deepfake detection has shown remarkable advances \cite{tolosana2020deepfakes, mirsky2021creation} in recent years. However, the emerging threat~\cite{deng2020reference}, which can precisely edit facial components partially, may fail these detection methods which focus on the entire face forgery detection. More importantly, for this brand-new forgery manner~\cite{deng2020reference}, we want to know where the falsification takes place.  Compared to the traditional deepfake forgeries, locating and segmenting fake facial components is increasingly important but not extensively noticed by research communities.

    In this paper, to make detection results more explainable, we focus on not only facial forgery detection but also fine-grained falsified facial components segmentation. We propose a collaborative feature learning approach to simultaneously detect manipulation and segment the falsified components. Our architecture requires a shared encoder between two tasks to enhance the latent features, which is followed by two decoders for implementing facial forgery detection and segmentation. With the help of face component editing method proposed in~\cite{deng2020reference}, we provide accurate pixel-level ground-truth of both entire or partial face manipulation, which is different from~\cite{dang2020detection} where the ground-truth is estimated. This facilitates our segmentation task. With the collaborative learning, detection and segmentation can boost each other efficiently. The main contributions are summarized as follows:
    
        \noindent (a) We introduce a segmentation problem for fine-grained face forgery detection. We propose a collaborative feature learning approach for not only facial manipulation detection but also segmentation targeting on SOTA deepfakes including generated by face swapping or facial components changing.
        
        \noindent (b) We construct a face forgery dataset including entire and partial face manipulation with pixel-level ground-truth. The dataset includes $30,000$ falsified face images generated by editing one or multiple facial components as shown in Fig.~\ref{fig:dataset-examples}
        
        \noindent (c) Our method achieves state-of-the-art performance. The mutual promotion between forgery detection and manipulated region segmentation is proved. The visualization results have shown that our proposed model always captures the artifacts on facial regions,  which is more reasonable. 
    
\section{Related work}
    
\subsection{Deepfake Generation}
    Deepfakes have attracted extensive researchers' attention. With the rapid development of deep learning, various forgery methods are proposed to generate compelling fake faces, including entire and partial facial forgeries.
    
    The entire facial forgery methods replace the whole face region with a forged face, such as \textit{Deepfakes} \cite{Deepfakes}, \textit{FS-GAN} \cite{nirkin2019fsgan}, and \textit{Face2Face} \cite{thies2016face2face}.   \textit{Deepfakes} trains two autoencoders with a shared encoder for reconstructing faces from two specific individuals. A forged face is generated from a source face by the trained encoder and the target decoder. Poisson image editing is leveraged to blend the fake face with the target image \cite{rossler2019faceforensics++, perez2003poisson}. 
    \textit{FS-GAN} is subject agnostic, which aims to remove the constraints of only swapping faces between two specific individuals. It introduces face segmentation and reenactment to adjust for both pose and expression variations. 
    Unlike the above approaches, \textit{Face2Face} attempts to edit the facial expressions in videos. Manual key-frame selection from two video input streams is implemented to generate a dense reconstruction of the face. 
    
    The partial facial forgery methods only tamper some specific facial regions. Previous work targets on changing some predefined attributes, such as the hair color \cite{choi2018stargan} and the age \cite{liu2019attribute}, or editing attributes by shape changes \cite{he2019attgan}. However, these methods mainly manipulate the faces by re-generating the entire images. Another noticeable category of methods has the ability of diverse and controllable facial component editing, e.g. \textit{R-Face} \cite{deng2020reference}. \textit{R-Face} takes an image inpainting model as the backbone, utilizing reference images as conditions for controlling the shape of face components. In the manipulation process, it maintains the other parts unchanged.

\subsection{Manipulation Detection and Segmentation}
    Due to the continuous development of deepfake generation techniques, various detection methods have also been proposed by the researchers to capture the flaws of deepfakes. The flaws of deepfakes are roughly divided into two categories, including pixel-level artifacts and semantically meaningful artifacts \cite{agarwal2020detecting}. 
    
    The deepfake detection methods based on the pixel-level artifacts treat this problem as a classification task and utilize a CNN to tackle it, such as \textit{MesoNet} \cite{afchar2018mesonet}, \textit{XceptionNet} \cite{Chollet_2017_CVPR, rossler2019faceforensics++}, and \textit{EffcientNet} \cite{tan2019efficientnet}. \textit{MesoNet} adopts a neural network with a small number of layers, focusing more on mesoscopic level features. \textit{XceptionNet}, as a modified version of Inception v3 \cite{szegedy2016rethinking}, has good detection performance on FaceForensics++ Dataset \cite{rossler2019faceforensics++}. Compared to \textit{XceptionNet}, \textit{EfficientNet} shows approximate or better performance on deepfake detection, which is widely adopted by Deepfake Detection Challenge solutions~\cite{dolhansky2020deepfake}. Semantically meaningful artifacts includes the deficiency of details in eyes and teeth \cite{8638330}, the incompatibility of appearance and behavior \cite{agarwal2020detecting}, inconsistent 3D head poses \cite{yang2019exposing} and blending boundaries \cite{li2020face}. The authors of \cite{yang2019exposing} observe that the splicing in deep fakes introduces the inconsistency between 3D head poses estimated from the facial landmarks and the central facial region. Face X-Ray \cite{li2020face} leverages the blending boundaries, which is a shared step in the deepfake generation process, for deepfake detection.
    
    Another important topic in deepfake detection is locating manipulated regions of deepfakes. Some methods for indicating the forged regions on deepfakes have been developed, such as utilizing class activation maps to predict manipulated regions \cite{li2019zooming},  Y-Shaped AutoEncoder \cite{nguyen2019multi}, and XceptionNet with Attention Map \cite{dang2020detection}. In Y-Shaped AutoEncoder~\cite{nguyen2019multi}, the authors propose a multi-task network with the classification, segmentation, and reconstruction tasks. For the modified XceptionNet with Attention Map, the authors attempt to learn an attention map to help the classification network focus on the fake regions, which can improve the performance and explainability of deepfake detection. The attention map is further utilized to locate manipulated regions. In addition, some previous general-purpose splicing localization and segmentation methods are not limited to face images, such as Hybrid CNN-LSTM \cite{bappy2017exploiting}, MAG \cite{kniaz2019point} and ManTra-Net \cite{wu2019mantra}. For the first method, the motivation is to learn the boundary discrepancy between the manipulated and non-manipulated regions. For MAG, the authors attempt to simultaneously enhance the splicing model and the image splice detection model during the adversarial training. And ManTra-Net consists of Image Manipulation Trace Feature Extractor and Local Anomaly Detection Network. It detects forged pixels by identifying local anomalous features, which is not limited to specific forgery methods.
    
    In our work, we focus more on the facial forgery segmentation task of partially edited faces. We design a network architecture with collaborative feature learning for deepfake detection and manipulated region  segmentation. XceptionNet is introduced into our architecture for improving feature learning capability. Finally, we show that the segmentation process contributes to enhance the detection interpretability.
    
\section{Method}
    In this section, we first detail the forgery generation process and give a definition of manipulation mask for our problem. Then we describe the proposed collaborative feature learning network for manipulation detection and segmentation. Finally, we explore the appropriate network for extracting pixel-level artifact features to improve forgery segmentation performance.
    
\begin{figure}
    \centering
    \includegraphics[width=0.45\textwidth]{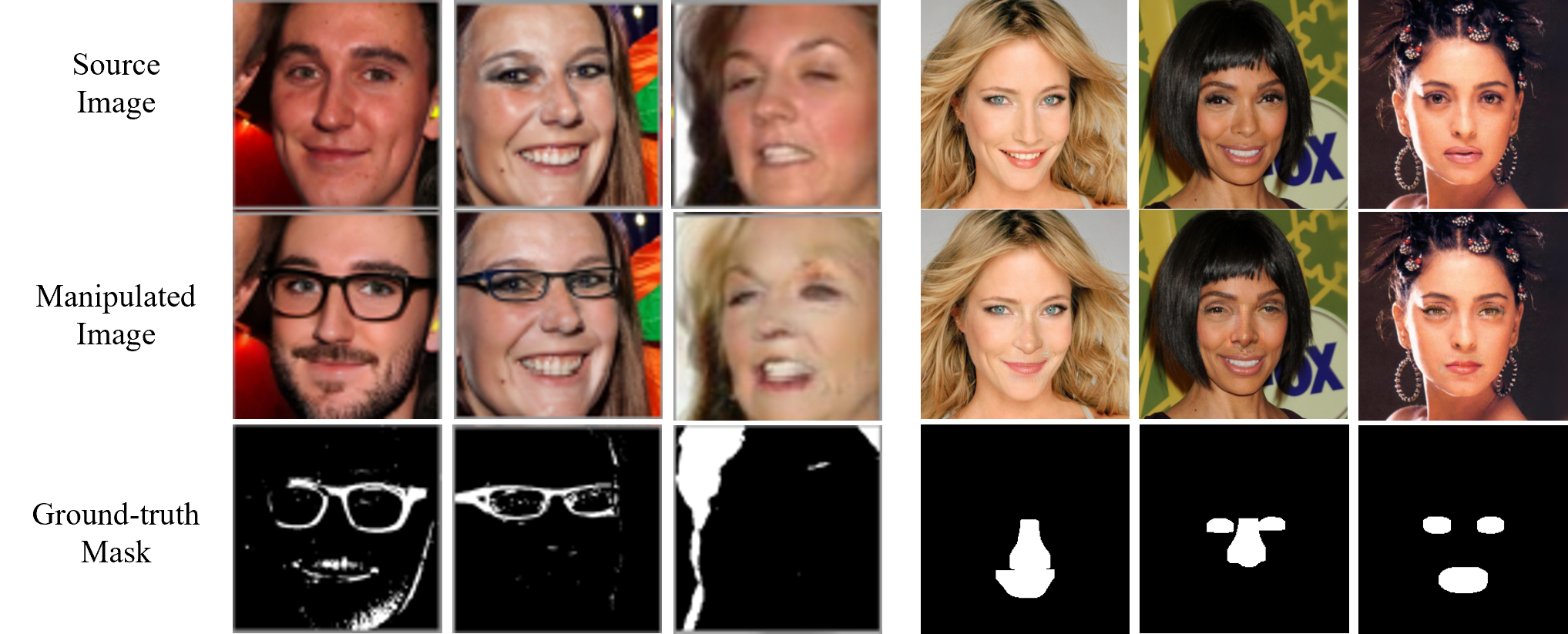}
    \caption{The comparison between manipulation masks in the previous work and ours. The left three columns are from \cite{dang2020detection}, while the right three columns are generated by R-Face \cite{deng2020reference}.}
    \label{fig:forgery-mask}
\end{figure}

\subsection{Facial Manipulation} 
    In this paper, we mainly focus on the precise localization and segmentation of falsified regions in manipulated face images, especially partially edited faces, whereas traditional forgery detection methods mainly output entire face as fake. For the emerging face manipulation method, \textit{R-Face}~\cite{deng2020reference}, which is able to precisely edit important facial components (e.g. eyes, noise and mouth), traditional detection may be failed, let alone segmentation. To this aim, we define the facial manipulation process as follows. 
    \begin{equation}
        I_{g} = G(I_{s}, M, I_{t}),
    \end{equation}
    where, $G(\cdot)$ is the generation process which requires three inputs, a source image $I_{s}$, a target image $I_{t}$, and the mask of the targeted facial components $M$. $I_{g}$ is the generated fake image. For entire face manipulation, $M$ is always the detected face region like Deepfakes~\cite{Deepfakes} or convex hull area of the face landmarks like FaceSwap. For facial components manipulation, $M$ is the corresponding facial components mask, as shown in Fig.~\ref{fig:forgery-mask}. Taking \textit{R-Face} for example, it utilizes source images as conditions for controlling the shape of target facial components. This method first occludes the facial components of the target image with $M$, and then conducts face in-painting with corresponding components from $I_{g}$. The generated partial manipulated face image can be defined as
    \begin{equation}
        I_{\hat{g}} = M * I_{g} + (1-M) * I_{t},
    \end{equation}
    Here, $*$ is an element-wise multiplication operator. Different from previous work \cite{dang2020detection}, which defines the difference between a fake image and its corresponding genuine image as the manipulation mask, we use $M$ as the ground-truth manipulation mask, as shown in Fig.~\ref{fig:forgery-mask}.
    
    Our main task is to predict the manipulation mask $M$ from a given image $I$ by a segmentation process. In addition, we also want to know whether $I$ is falsified or not. Therefore, besides manipulated region  segmentation, we also output detection results from our model $F(\cdot)$. The joint task is expressed as:
    \begin{equation}
        \{p, S\} = F(I),
    \end{equation}
    where $p$ is the probability that the suspected image is falsified, and $S$ is the estimated manipulation mask.

\begin{figure}
    \centering
    \includegraphics[width=0.45\textwidth]{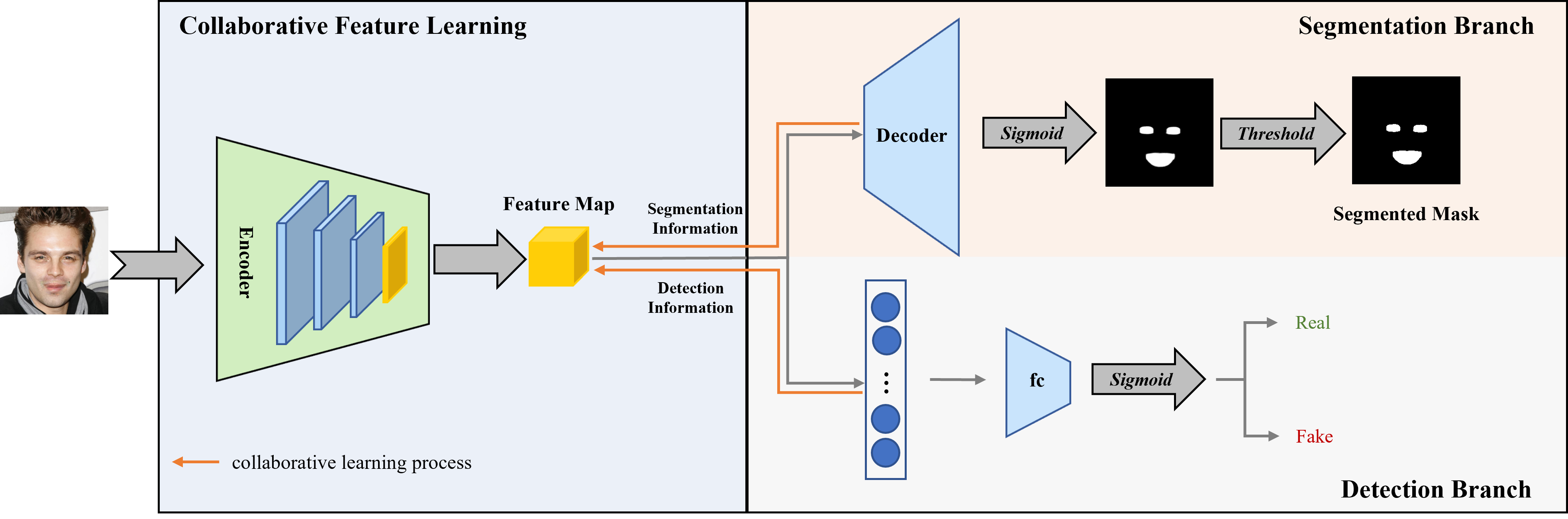}
    \caption{The proposed network for manipulation detection and segmentation.}
    \label{fig:architecture}
\end{figure}
\subsection{Collaborative Forgery Feature Learning}
    As mentioned above, our objective is to precisely segment forged regions and detect manipulation. Instead of separately performing the two tasks, we resort to collaborative feature learning for both them to improve their performance. On the one hand, facial manipulation segmentation provides the manipulated regions, which can help detector focus on these regions for more precise prediction. On the other hand, manipulation detection helps segmentation correct the segmented mask. If the image is genuine, the segmented mask should be predicted as zero.
    
    In the training phase, the two tasks simultaneously influence the feature extraction process. In this process, we aim to optimize the feature representation for improving the performance of the two tasks. Specifically, for locating and segmenting manipulated regions, an intuition is setting this problem as a pixel-level segmentation task. Given the ground-truth of the manipulation mask $M \in \{0, 1\}^{h\times w}$, we use BCE (Binary Cross-Entropy) loss as the segmentation loss to measure the difference between the predicted segmentation mask $S \in (0, 1)^{h \times w}$ and $M$:
    \begin{equation}
        L_{seg} = - \frac{1}{N} \sum_{k=1}^N \sum_{i=1}^h \sum_{j=1}^w L_p(s_{k,i,j}, m_{k,i,j}),
    \label{loss:Lseq}
    \end{equation}
    where $N$ is the number of the forgery samples, $s_{k,i,j}$ and $m_{k,i,j}$ are respectively the pixel with the coordinate $(i, j)$ in the $S$ and $M$ of $k$-th sample, and $L_p$ is the BCE loss:
    \begin{equation}
        L_p(s, m) = \frac{(1-m)log(1-s) + mlog(s)}{h \times w},
    \label{loss:Lp}
    \end{equation}
    The predicted mask is obtained by a further binarization on the segmentation output.
    
    The manipulation detection branch shares the encoder with the segmentation branch. The fully connected layers are connected to the encoder for classification. The output of \textit{fc} layer is normalized into the range of $(0, 1)$ by $Sigmoid$ function. We also use BCE loss as the detection loss $L_{det}$:
    \begin{equation}
        L_{det} = - \frac{1}{N} \sum_{k=1}^N (1-y_k)log(1-p_k) + y_klog(p_k),
    \label{loss:Ldet}
    \end{equation}
    Here, $y_k \in \{0, 1\}$ is the lable of the $k$th sample, and $p_k \in (0, 1)$ is the prediction output. 
    
    The total loss is the weighted sum of the two losses:
    \begin{equation}
        L = \lambda_{det}L_{det} + \lambda_{seg}L_{seg}.
    \label{loss:all}
    \end{equation}
    Here, both $\lambda_{seg}$ and $\lambda_{det}$ are set to $1$ empirically.

\subsection{Network Architecture}
\label{sec:network}
    As shown in Fig.~\ref{fig:architecture}, we design an architecture to perform collaborative feature learning for facial manipulation detection and segmentation by sharing a feature encoder. In the facial manipulated region segmentation task, the extracted feature maps are fed to its corresponding decoder, followed by a $Sigmoid$ function to map the output into $(0, 1)$. The output is a single-channel map with the same size as the input image. Every pixel of the output map predicts the probability that the corresponding pixel is alerted. Finally, the output is binarized to obtain the segmentation mask. 
    
    For the manipulation detection branch, the extracted features are fed to fully connected (\textit{fc}) layers after performing $\textit{Global Average Pooling}$. The final output is the probability that the image is falsified.
    
    We use XceptionNet \cite{Chollet_2017_CVPR}, as the shared feature extractor where the last $fc$ layer is used for feature representation, due to its good performance on the forgery detection task~\cite{rossler2019faceforensics++, dolhansky2019deepfake}. The segmentation decoder is built by a de-convolution network which is inspired by DCGAN~\cite{radford2015unsupervised}.

\section{Experiments}
\subsection{Dataset}
    In our task, we aim to locate and segment the manipulated regions of face images. However, most previous datasets only comprise of the forgeries that the entire faces are generated or manipulated. Therefore, we create a dataset for our task. Our dataset contains entirely and partially manipulated faces (e.g. eyes, nose, and mouth). 
    
    \textbf{Data Collection.} As mentioned above, we care about both the entirely and partially manipulated faces.
    
    \textit{Entirely manipulated faces.} We utilize a part of video clips from FaceForensics++ \cite{rossler2019faceforensics++}, which contains $1,000$ real videos and the corresponding $4,000$ fake videos generated by different forgery methods. We utilize the deepfakes generated by $FaceSwap$ and $Deepfakes$ and their corresponding mask data. The $RetinaFace$ \cite{deng2019retinaface} is further employed to extract the facial regions from the video frames. The side length of the detected face bounding boxes are enlarged by a factor of $1.3$ for obtaining the full face images.
    
    \textit{Partially manipulated faces.} We utilize \textit{R-Face} \cite{deng2020reference} to edit facial components (e.g. eyes, nose, and mouth) while maintaining others unchanged, which can give us precise ground-truth manipulation masks. CelebAMask-HQ \cite{lee2020maskgan} dataset is employed to generate partially manipulated face images. We randomly select one or multi components to manipulate.
    
    \textit{Real face data.} Since the forgery data is generated from the real data of FaceForensics++ and CelebAMask-HQ dataset by different manipulation methods, we utilize them as our real face data.
    
    More details and examples of the constructed dataset are shown in Table~\ref{tab:dataset} and Fig.~\ref{fig:dataset-examples}. 
    \begin{table}[t]
        \centering
        \resizebox{84mm}{!}{
        \begin{tabular}{c c c c c c}
        \hline
           \multirow{3}{*}{Label} & \multicolumn{2}{c}{\multirow{2}{*}{Real}} & \multicolumn{3}{c}{Fake}\\
        \cline{4-6}
             & & & \multicolumn{2}{c}{EM} & PM \\
        \cmidrule(r){2-3} \cmidrule(r){4-5} \cmidrule(r){6-6}
            & FFpp-Ori & CAM-HQ & FS & Df & R\\
        \hline
        \# Video Clips & $1,000$ & - & $1,000$ & $1,000$ & - \\
        \# Images/Frames & $509,124$ & $30,000$ & $405,401$ & $509,086$ & $30,000$ \\
        \hline
        \end{tabular}}
        \caption{Statistics of our dataset. EM and PM respectively denote the entirely and partially manipulated faces. FFpp-Ori, FS and Df are the real and forgery images generated by $FaceSwap$ and $Deepfakes$ in FaceForensics++ dataset, respectively. CAM-HQ is CelebAMask-HQ dataset. R is the paritially manipulated facial images generated by \textit{R-Face}.}
        \label{tab:dataset}
    \end{table}
    
    \textbf{Data Splitting.} We repeat the official procedure to split the training, validation, and testing dataset of FaceForensics++. To balance the size of different forgery and real data, we design a sampling strategy.  We randomly sample $60$ faces from every origin video, and randomly select $30$ forged faces from every fake video. We totally sample $60,000$ real faces and $60,000$ forged faces generated by two manipulation methods. For the \textit{R-Face} and CelebAMask-HQ dataset, we adopt all of them in our experiments. We refer to the sorted list in \cite{deng2020reference} to split CelebAMask-HQ and \textit{R-Face} dataset. The first $27,000$ faces are set as training data, the last $1,500$ faces are set as testing data, and the other faces are used as validation data. The statistics of the data employed in our experiments are detailed in Table~\ref{tab:data-statistics}.
    
    \begin{table}[t]
        \centering
        \resizebox{84mm}{!}{
        \begin{tabular}{c c c c c c c c}
        \hline
           \multirow{2}{*}{} & \multicolumn{2}{c}{Real} & \multirow{2}{*}{All Real} & \multicolumn{3}{c}{Fake} & \multirow{2}{*}{All Fake}\\
        \cmidrule(r){2-3} \cmidrule(r){5-7}
           & FFpp-Ori & CAM-HQ &  & FS & Df & R & \\
        \hline
        \# Train & $43,200$ & $27,000$ & $70,200$ & $24,600$ & $24,600$ & $27,000$ & $70,200$\\
        \# Val & $8,400$ & $1,500$ & $9,900$ & $4,200$ & $4,200$ & $1,500$ & $9,900$\\
        \# Test & $8,400$ & $1,500$ & $9,900$ & $4,200$ & $4,200$ & $1,500$ & $9,900$\\
        \hline
        \end{tabular}}
        \caption{Statistics of data splitting.}
        \label{tab:data-statistics} 
    \end{table}

\subsection{Experiment Setup}
    In this paper, we mainly focus on detection and segmentation of facial manipulation. Therefore, in all experiments, we respectively report the detection accuracy (Acc) and intersection over union (IoU) for evaluating the detection and facial manipulated region segmentation performance. We also report more detailed performance for the two tasks, as described in the following experiments.

\begin{figure}[t]
    \centering
    \includegraphics[width=0.4\textwidth]{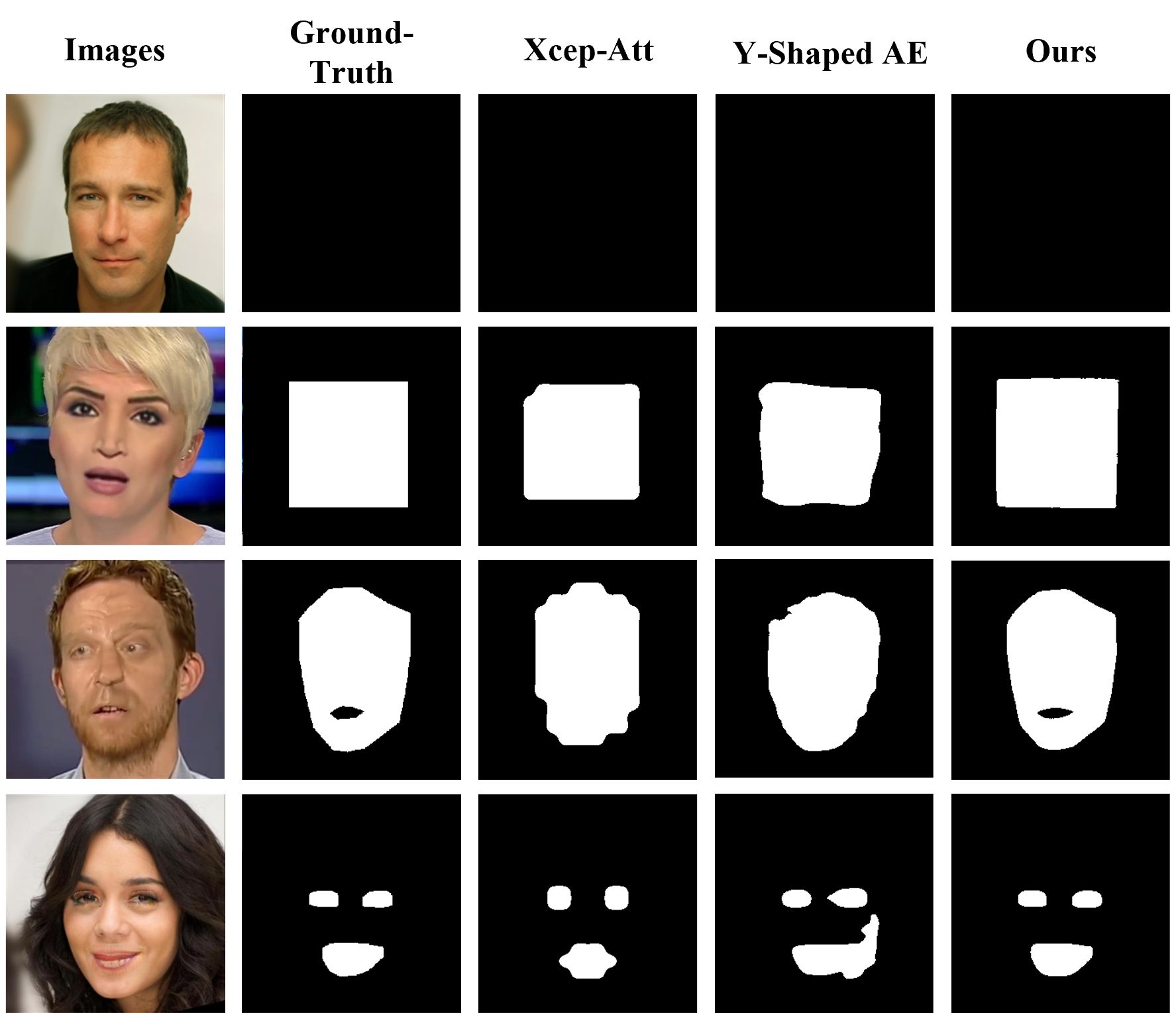}
    \caption{The visualization of the facial manipulated region  segmentation results of different methods. The left two columns are the facial images and the corresponding ground-truth. The right three columns are respectively the predicted masks by Xcep-Att \textit{reg}, Y-Shaped AE and our method (Xcep-Multi).}
    \label{fig:localization-results}
\end{figure}

\begin{table*}[t]
    \centering
    \resizebox{170mm}{!}{
    	\begin{tabular}{c | c| c c c c c | c| c c c c c}
    	\hline
    	  \multirow{2}{*}{Method} & \multicolumn{6}{c|}{Detection} & \multicolumn{6}{c}{Segmentation} \\
    	\cline{2-13}
    	    & Acc-All & Acc-Real & Acc-Fake & Acc-FS & Acc-Df & Acc-R & IoU-All & IoU-Real & IoU-Fake & IoU-FS & IoU-Df & IoU-R\\
    	\hline
    	    XceptionNet & 0.9861 & 0.9869 & \textbf{0.9853} & 0.9824 & 0.9867 & \textbf{0.9893}  & -- & -- & -- & -- & -- & --\\
    	    MesoInception-4 & 0.7884 & 0.9704 & 0.6064 & 0.4864 & 0.5952 & 0.9733 & -- & -- & -- & -- & -- & --\\
	    \hline
    	    U-Net & -- & -- & -- & -- & -- & -- & 0.8678 & 0.8098 & 0.9258 & 0.9105 & 0.9292 & 0.9592 \\
    	    ManTra-Net & -- & -- & -- & -- & -- & -- & 0.8271 & 0.8273 & 0.7823 & 0.7681 &  0.7332 & \textbf{0.9594} \\
    	\hline
       \specialrule{0em}{2pt}{2pt}
       \hline
    	    Y-Shaped AE & 0.8440 & 0.7079 & 0.9801 & 0.9581 & \textbf{0.9990} & 0.9887 & 0.9224 & \textbf{1.0000} & 0.8448 & 0.8460 & 0.8546 & 0.8140 \\
    	    Xcep-Att \textit{reg} & 0.9671 & 0.9638 & 0.9704 & 0.9595 & 0.9748 & 0.9887 & 0.8655 & 0.9098 & 0.8213 & 0.8331 & 0.8442 & 0.7239 \\
    	\hline
    	    \textbf{Ours}  & \textbf{0.9910} & \textbf{0.9982} & \textbf{0.9837} & \textbf{0.9871} & 0.9783 & \textbf{0.9893} & \textbf{0.9659} & 0.9944 & \textbf{0.9373} & \textbf{0.9457} & \textbf{0.9336} & \textbf{0.9239} \\
    	\hline
    	\end{tabular}
    	}
    \caption{Comparison with some previous related methods.}
    \label{tab:comparison-with-others}
\end{table*}

\subsection{Results on Manipulation Detection and Segmentation}
    In this part, we compare our proposed method with previous work on manipulation detection and segmentation. Different from the previous facial manipulation segmentation tasks, we attempt to segment the manipulated regions both in the entirely and partially manipulated faces. We implement all of the methods under the same experiment settings for a fair comparison.
    
    As shown in Table~\ref{tab:comparison-with-others}, we respectively implement the previous methods for detection-only, segmentation-only, and both. All of the methods are re-trained in our training dataset and tested in testing dataset. For the manipulation detection, we set XceptionNet \cite{Chollet_2017_CVPR} and MesoNet (MesoInception-4) \cite{afchar2018mesonet} as the comparison methods. For manipulation segmentation task, besides U-Net \cite{ronneberger2015u}, we also employ a forensics method, ManTra-Net \cite{wu2019mantra}. These methods provide us the baselines of detection and segmentation tasks. For the methods simultaneously tackling these two problems, we select Y-Shaped Autoencoder (Y-Shaped AE) \cite{nguyen2019multi} and XceptionNet with Attention Mechanism (Xcep-Att) \cite{dang2020detection}. For Xcep-Att, we merely utilize the regression-based method to estimate the attention maps (Xcep-Att \textit{reg}), which obtains the best performance in \cite{dang2020detection}.  Table~\ref{tab:comparison-with-others} shows the performance of all methods. 
    From the table, we can find that our method can achieve  better performance in most cases on the two tasks. 
    
    Compared to the detection-only and segmentation-only methods, our method shares features between two tasks by collaborative feature learning and achieves better performance, as shown in Table~\ref{tab:comparison-with-others}. An interesting phenomenon is that the $IoU$ score of U-Net and ManTraNet on partially edited faces obviously outperforms that on entirely manipulated faces. This may be because that the incorrectly segmented regions on fine-grained manipulated images are smaller than those on face swapping deepfakes.
    Additionally, Table~\ref{tab:comparison-with-others} shows that compared to previous multi-task methods, our method also achieves better results. Our method predicts pixel-level manipulation masks for precise segmentation, instead of resizing the attention maps like Xcep-Att. Besides, reconstructing the input images in Y-Shaped AE may introduce additional information, which influences the segmentation process, especially on the fine-grained falsified faces. 
    The visualization of the manipulation segmentation results is shown in Fig.~\ref{fig:localization-results}.
\subsection{Ablation Study}
    \begin{table}[t]
        \centering
        \resizebox{84mm}{!}{
        	\begin{tabular}{c | c c c c c | c}
        	\hline
        	    & Acc-Real & Acc-Fake & Acc-FS & Acc-Df & Acc-R & Acc-All \\
        	\hline
        	  Ours (U-Net) &  0.9895 & 0.9830 & 0.9810 & \textbf{0.9829} & \textbf{0.9893} & 0.9863 \\
        	  Ours (Xcep) & \textbf{0.9982} & \textbf{0.9837} &  \textbf{0.9871} & 0.9783 & \textbf{0.9893} & \textbf{0.9910} \\
        	\hline
        	\hline
        	    & IoU-Real & IoU-Fake & IoU-FS & IoU-Df & IoU-R & IoU-All \\
        	\hline
        	  Ours (U-Net) &  0.7438 & 0.9213 & 0.9172 & 0.9164 & \textbf{0.9464} & 0.8326 \\
        	  Ours (Xcep) & \textbf{0.9944} & \textbf{0.9373} &  \textbf{0.9457} & \textbf{0.9336} & 0.9239 & \textbf{0.9659} \\
        	\hline
        	\end{tabular}
        	}
        \caption{Our architecture based on two feature extractor networks.}
        \label{tab:extractor-comparison}
    \end{table}
    
    \textbf{Effect of Feature Extractor Networks.} We try to investigate the effect of different feature extractor networks for more details. For comparison, we use U-Net based architecture as an alternative. It is noticed that we cancel the connection between the encoder and decoder of U-Net in our architecture. This change is intended to reduce the extra information transfer from the encoder to the decoder, which can force the extracted features to integrate the information of manipulated regions. The $fc$ layers are stacked by two fully connected layers and a $ReLU$ activation function for outputting the detection results. This can be considered as the baseline. Table~\ref{tab:extractor-comparison} shows the comparison results of XceptionNet and U-Net based networks. We can find that XceptionNet based architecture does improve the performance of the detection and segmentation. This is because, for the shared feature extractor, a larger and deeper network (XceptionNet) can extract more powerful features. The experiments demonstrate the superiority of XceptionNet as the shared feature encoder.

\begin{table}[t]
    \centering
    \resizebox{84mm}{!}{
    	\begin{tabular}{c | c c c c c | c}
    	\hline
    	    & Acc-Real & Acc-Fake & Acc-FS & Acc-Df & Acc-R &  Acc-All\\
    	\hline
    	  Ours-\textit{No Seg} & 0.9869 & 0.9853 & 0.9824 & \textbf{0.9867} & \textbf{0.9893} & 0.9861\\
    	  Ours  & \textbf{0.9982} & \textbf{0.9837} & \textbf{0.9871} & 0.9783 & \textbf{0.9893} & \textbf{0.9910}\\
    	\hline
    	\hline
    	    & IoU-Real & IoU-Fake & IoU-FS & IoU-Df & IoU-R & IoU-All\\
    	\hline
    	  Ours-\textit{No Det} & 0.9886 & 0.9310 & 0.9362 & 0.9282 & \textbf{0.9244} & 0.9598\\
    	  Ours  & \textbf{0.9944} & \textbf{0.9373} & \textbf{0.9457} & \textbf{0.9336} & 0.9239 & \textbf{0.9659}\\
    	\hline
    	\end{tabular}
    	}
    \caption{Comparison between the results of our architecture with and without collaborative feature learning.}
    \label{tab:information-interaction}
\end{table}

    \textbf{Benefit of Collaborative Feature Learning.} For exploring the benefit of collaborative feature learning, we further elaborate the comparison results of our architecture with and without it. Specifically, we respectively cut off the detection and segmentation branches in our architecture. As shown in Table~\ref{tab:information-interaction}, \textit{No Seg/Det} denotes without the segmentation/detection branch from our architecture. Without the collaborative feature learning, we set the two results as the baseline of our method. The comparison experiments show that compared to the baseline, our method with the collaborative feature learning has better or comparable performance in the two tasks. This demonstrates that the collaborative feature learning for the two tasks does improve the performance of each other.

\begin{figure}[t]
    \centering
    \includegraphics[width=0.4\textwidth]{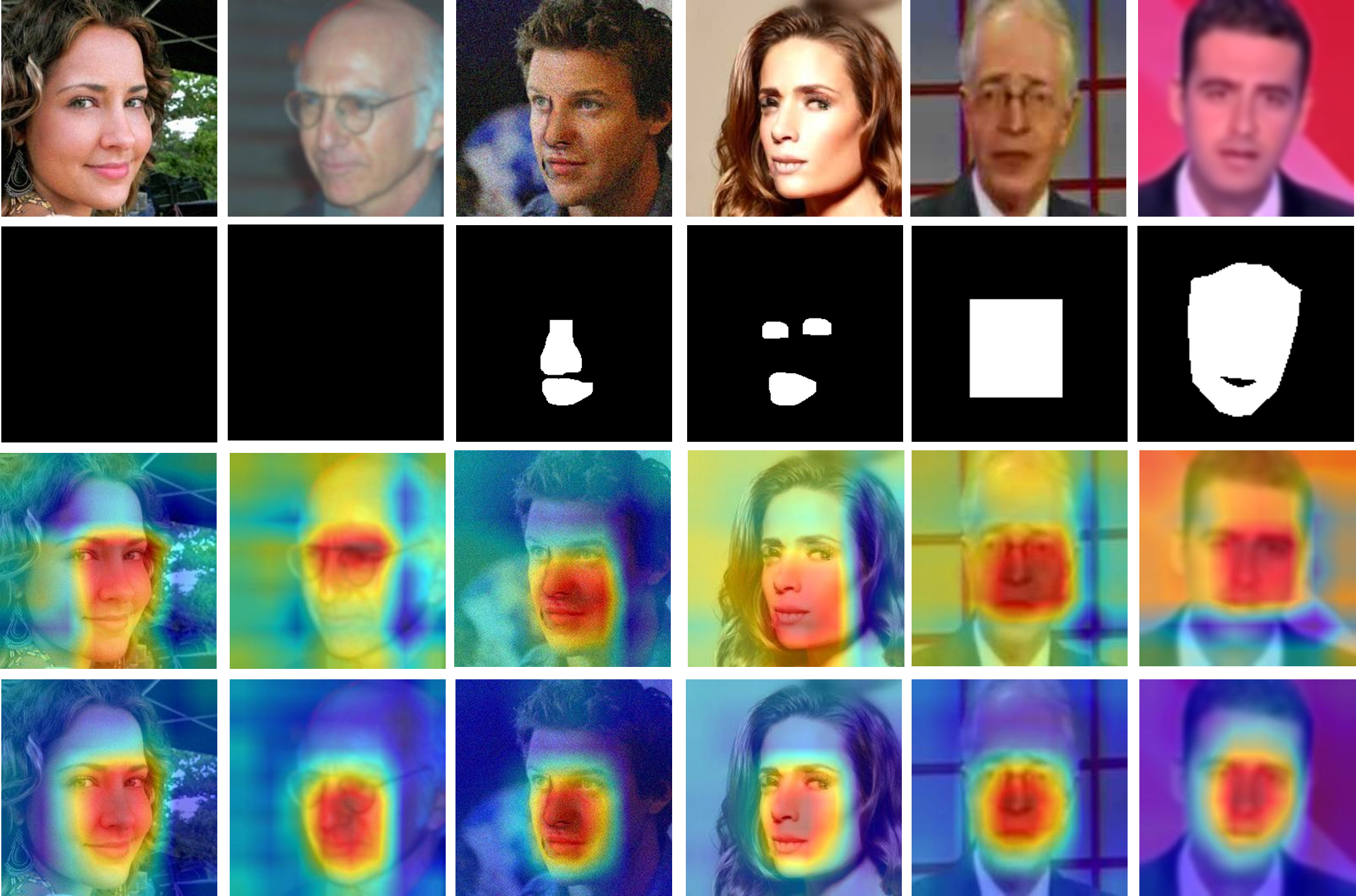}
    \caption{The results of the improving interpretability. The top two rows are facial images and the corresponding fake masks. The third and fourth rows are the class activation maps of XceptionNet and our method in the deepfake detection task.}
    \label{fig:interpretability}
\end{figure}

\subsection{Better Interpretability}
	In our method, the detection and segmentation branches have a shared encoder. The feature extraction process is simultaneously influenced by each other. As the previous experiments showing, the segmentation task resorts to the good feature extraction capability of the detection process for improving its performance. Conversely, the segmentation process also provides manipulation evidence to the detection task.
	
	Therefore, we employ Grad-CAM++ \cite{chattopadhay2018grad} to show the class activation map of the last layer in the shared encoder, as shown in Fig.~\ref{fig:interpretability}. We compare our method with XceptionNet. Compared to the original XceptionNet, our method forces the detection process to focus more on the facial regions rather than backgrounds or surroundings where manipulation seldom occurs, which is more convincing for explaining the manipulation detection. A reasonable explanation is that segmentation process helps detection process concentrate on all of the suspected regions. The suspected regions should include all facial parts that are probably  manipulated. In our task, all of the facial regions could be falsified. Hence, our method focuses on the entire face region as the suspected regions both in real and fake images for especially further investigation.
	
\section{Conclusion}
    In this work, we introduce a segmentation problem for fine-grained face forgery detection. Our proposed method leverages collaborative feature learning to simultaneously tackle facial forgery detection and segmentation. We construct a face forgery dataset, including entire and partial face manipulation with pixel-level ground-truth, to facilitate the proposed tasks. We empirically show that the usage of collaborative feature learning improves the performance of manipulation segmentation and detection mutually and enhances the interpretability of forgery detection. Finally, Our method achieves SOTA performance in comparison to previous methods on both two tasks.

\section{Acknowledge}
    Dr. Qiyao Deng (dengqiyao@cripac.ia.ac.cn) provides important assistance in our dataset construction. We would like to give special thanks here. 
    
    This work is supported by the National Key Research and Development Program of China under Grant No. 2020AAA0140003 and the National Natural Science Foundation of China (NSFC) under Grants 61972395, U19B2038.
{\small
\bibliographystyle{ieee}
\bibliography{main}
}

\end{document}